\documentclass[10pt,twocolumn,letterpaper]{article}
\usepackage[pagebackref,breaklinks,colorlinks]{hyperref}
\usepackage[algorithms]{wacv}
\usepackage{amsfonts,amssymb,amsmath,dsfont}
\usepackage{microtype}
\usepackage{graphicx}
\usepackage{booktabs}
\usepackage{hyperref}
\usepackage{multirow}
\usepackage{multirow}
\usepackage{makecell}
\usepackage{amsmath}
\usepackage{amssymb}
\usepackage{mathtools}
\usepackage{amsthm}
\usepackage{arydshln}
\usepackage{xcolor}
\usepackage{color}
\usepackage{algorithm}
\usepackage[para,online,flushleft]{threeparttable}
\usepackage[square,sort,comma,numbers]{natbib}
\usepackage{caption}
\usepackage{times}
\usepackage{epsfig}
\usepackage{graphicx}
\usepackage{booktabs}
\usepackage{pifont}
\usepackage{tabularx}
\usepackage[textsize=tiny]{todonotes}

\newcommand{\method}[1]{\ifthenelse{\equal{#1}{full}}{Primary-Auxiliary Objectives Association}{PAOA}}
\graphicspath{{./figs/}{./figs/result/}}

\newcommand{\cmark}{\ding{51}}
\newcommand{\xmark}{\ding{55}}
\newcommand{\best}[1]{{\color{red}\textbf{#1}}}
\newcommand{\scnd}[1]{{\color{blue}\underline{#1}}}

\graphicspath{{./figs/}{./figs/result/}}

\def\ie{\emph{i.e.}, }

\def\eg{\emph{e.g.}, }

\def\wacvPaperID{337}
\def\confName{WACV}
\def\confYear{2024}

\begin{document}

\title{Mitigate Domain Shift by \method{full} \\ for Generalizing Person ReID}

\author{
Qilei Li,
Shaogang Gong
\\
\small {Queen Mary University of London}\\
\tt\small \{q.li, s.gong\}@qmul.ac.uk
}

\maketitle

\begin{abstract}
      While deep learning has significantly improved ReID model
      accuracy under the independent and identical distribution (IID) assumption,
      it has also become clear that such models degrade
      notably when applied to an unseen novel domain due to unpredictable/unknown domain shift.
      Contemporary domain generalization (DG) ReID models
      struggle in learning domain-invariant representation
      solely through training on an instance classification objective.
      We consider that a deep learning model is heavily influenced
      and therefore biased towards domain-specific characteristics,
      \eg background clutter, scale and viewpoint variations,
      limiting the generalizability of the learned model,
      and hypothesize that the pedestrians
      are domain invariant
      owning they share the same structural characteristics.
      To enable the ReID model to be less domain-specific
      from these pure pedestrians,
      we introduce a method that guides model learning
      of the primary ReID instance classification objective by a concurrent auxiliary learning objective on
      weakly labeled pedestrian saliency detection.
      To solve the problem of conflicting optimization criteria in
      the model parameter space between the two learning objectives,
      we introduce a \method{full} (\method{abbr}) mechanism
      to calibrate the loss gradients of the auxiliary task
      towards the primary learning task gradients.
      Benefiting from the harmonious multitask learning design, 
      our model can be extended with the recent test-time diagram
      to form the \method{abbr}+,
      which performs on-the-fly
      optimization against the auxiliary objective 
      in order to maximize the model's generative
      capacity in the test target domain.
      Experiments demonstrate the superiority of the proposed \method{abbr} model.
      \end{abstract}
\section{Introduction}
Person Re-IDentification (ReID) \cite{li2018harmonious,zheng2019joint,zhang2020relation,li2021local}
is a fundamental task
which aims to retrieve the same pedestrian
across non-overlapping camera views
by measuring the distances among representations of all the candidates
in a pre-learned discriminative feature space.
However,
like most deep-learning models,
current ReID techniques are built based on an intrinsic assumption of
independent and identical distribution (IID) between training and
test data. The IID assumption becomes mostly invalid across different domains when training and
test data are not from the same environment.
As a result, most contemporary ReID models suffer from dramatic degradation
when applied to a new domain
~\cite{luo2020generalizing,choi2021meta,wei2018person}.
Domain Generalization (DG) methods
\cite{zhou2021domain,zhou2020learning,mahajan2021domain},
which aim to learn a generalizable model between a source and a
target domain have been explored by recent studies to address this problem.

\begin{figure}[h]
      \centering
      \includegraphics[width=\linewidth]{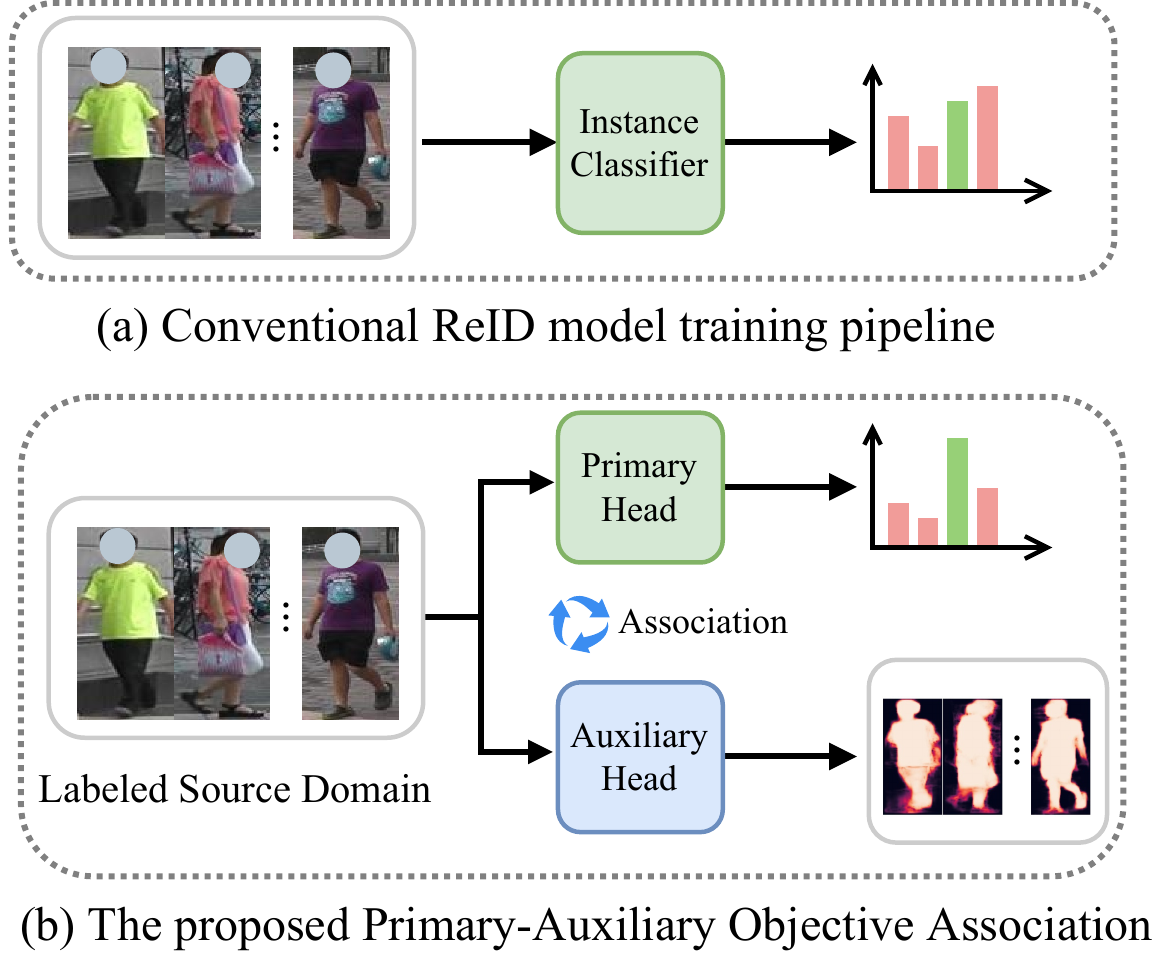}
      \caption{Comparing a standard Domain Generalization ReID
            model and the proposed \textit{\protect\method{full}} (\protect\method{abbr}) model.
            A DG model is typically trained by optimizing an instance classification objective,
            which can suffer from overfitting to domain-specific
            characteristics, \eg luminance, background, scale, and
            viewpoint. The \protect\method{abbr} model considers learning
            jointly a weakly labeled/supervised auxiliary saliency detection
            task concurrently with the primary task of the discriminative person
            ReID. This is achieved by calibrating
            the gradient of the auxiliary task against that of the primary objective as its reference.
      }
      \label{fig:method_compare}
      \vspace{-1em}
\end{figure}

A number of DG ReID methods have been developed
to mitigate performance degradation
caused by domain shift
between training (source) data and test (target) data.
They can be broadly categorized into three main groups:
(1) Learning from diversified training samples~\cite{jin2020style,ang2021dex},
(2) Aligning the distribution of source domains
by data statistics~\cite{zhuang2020rethinking, zhuang2021camera, jia2019frustratingly},
(3) Exploiting meta-learning
\cite{choi2021meta,zhou2020learning,dai2021generalizable,zhao2021learning} 
to mimic source-target distribution discrepancies.
The first category confers advantages to a model through the utilization of a diversified training dataset
by either image sample augmentation or feature distribution expansion.
The second category aims to learn a source-invariant model
by aligning the training data,
and expecting it to be invariant
for the target domain.
The third category focuses on simulating the training/testing discrepancy.
Despite some performance improvement from these methods,
their overall
performances across domains remain poor,
\eg the latest SOTA models \cite{choi2021meta,zhao2021learning}
can only achieve below 20\% mAP on the MSMT17 benchmark.
This highlights the limitation of overfitting in the current DG ReID
models and their inability to learn a more generalizable cross-domain
model representation.
We consider this is due to the not-insignificant interference
of domain-specific contextual scene characteristics
such as background, viewpoint, and object distances to a camera (scale),
which are identity-irrelevant but can change significantly across
different domains. Contemporary DG ReID models are mostly trained
by an instance-wise classification objective function, 
indirectly learning person foreground attention selection
(Figure~\ref{fig:method_compare}(a)). They are sensitive to such
domain-specific but identity-irrelevant contextual information,
resulting in the misrepresentation of person foreground attention and
leading to less discriminative ReID representation.
This likely causes notable ReID performance degradation
on models trained and deployed in different domains.
To mitigate the impact of domain-specific contextual attributes, 
an intuitive solution is to isolate the pedestrian object
to acquire a domain-invariant representation.
Several endeavors~\cite{he2020guided,zhu2020identity,song2018mask} have been made
to guide the person identification network focusing on the pedestrian
with the human saliency prior,
which can point out the attentive region 
relevant to the human subject.
These methods have certain limitations, 
either relying on exhaustive manual masking~\cite{song2018mask} 
or lacking an appropriate training objective~\cite{he2020guided,zhu2020identity} 
to ensure the accuracy of the generated segmentation mask.
Besides this, it is crucial to note that 
these methods fail to consider the potential worst-case scenario 
in which the saliency attention prior may be inaccurate,
further leading to negative impacts on identification rather than improvement.

In this work,
we address this problem
by introducing a novel model learning regularization method called
\textit{\method{full}} (\method{abbr}).
Our aim is to minimize domain-specific contextual interference in model learning
by focusing more on the domain-invariant person's unique characteristics.
This is achieved by introducing the association of learning the
primary instance classification objective function with an auxiliary
weakly labeled/supervised pedestrian saliency detection objective function,
the idea is illustrated in Figure~\ref{fig:method_compare}(b).
Specifically,
\method{abbr} is realized in two parts:
(1) Additionally train a pedestrian saliency detection head
with an auxiliary supervision to assist in focusing the primary ReID
discriminative learning task on more domain-invariant feature characteristics.
(2) Eliminate the interference attributed to inaccurate saliency labels
by calibrating the gradients of the shared feature extractor raised from the
weakly-labeled auxiliary learning task towards that of the primary
task as a reference when they are in conflict~\cite{sener2018multi}.
This association mechanism helps ensure the ReID model learns
to attentively focus on generic
yet discriminative pedestrian information
whilst both learning tasks are harmoniously trained.

Our contributions are:
(1) We introduce the idea of optimizing a more domain-generic ReID
learning task 
that emphasizes domain-invariant pedestrian characteristics
by associating the ReID instance
discriminative learning objective to an auxiliary
pedestrian saliency detection objective 
in a way that does not create conflicts 
or hinder the effectiveness of primary objective.
(2) We formulate a novel regularization
called \method{full} (\method{abbr}) to implement the
proposed association learning.
It jointly trains the primary and auxiliary tasks
with referenced gradient calibration
to solve the conflicting optimization criteria
between the two learning objectives,
and promote the learning of a more domain-generic ReID model.
(3) We further explore the target domain test data characteristics
by incorporating the \method{abbr} regularization
into a deployment-time model online optimization
process. To that end, we formulate a \method{abbr}+ mechanism
for on-the-fly target-aware model optimization
and show its performance benefit.

\section{Related Work}
\vspace{-1em}
\noindent \paragraph{Domain Generalizable ReID} (DG ReID)
assuming the absence of target domains during training,
aims to learn a generalizable model
which can extract discriminative representations
in any new environment.
It's naturally challenging but practical
and has attracted increasing attention.
Contemporary studies typically fall into three primary classifications:
(1) To benefit the model from the diverse training data
achieved by augmentation.
(2) To align the target domain
with the BN statistics calculated over the source domain.
(3) To mimic the train/test discrepancy
with meta-learning.
Despite the improvement obtained by these SOTA models,
significant room for improvement remains,
as indicated by the low mAP scores,
\eg less than 20\% on MSMT17 and less than 40\% on CUHK03.
This is attributed to the domain-specific interference
in the source domain
that limits the learning of a domain-invariant model.
In this work,
we aim to tackle this issue
by guiding the model to focus on the discriminative pedestrian area
with the tailored auxiliary task,
and propose the \method{abbr} regularization for that end.

\vspace{1em}
\noindent \textbf{Salient Object Detection}~\cite{borji2019salient}
aims to identify objects or regions
that are visually more attentive than the surrounding areas.
It has been significantly boosted
solely by the rapid development of deep learning.
Current detection models are usually trained end-to-end
and output a fine-grained saliency map at the pixel level.
In this work,
we design the auxiliary task
with the pedestrian saliency detection objective.
Instead of exhaustively labeling the pedestrian area manually
as the previous work~\cite{song2018mask},
we propose to use weakly labeled data
generated by a trained salient object detection model,
to benefit from large-scale training.
The recent work GASM~\cite{he2020guided} shares the similar spirit to ours
by employing weakly labelled saliency masks as an additional prior.
However,
GASM simply trains the saliency detection layers 
with the classification network
while omitting the potential worst-case 
where the weak label is not accurate
and cause potential conflict optimization direction
during model training.
In contrast, 
our method focuses on the \textit{association} between 
instance classification and saliency detection objectives
by the proposed referenced gradient calibration mechanism, 
which promotes the learning of the primary objective while mitigating the conflicts between the primary and auxiliary tasks.

\vspace{1em}
\noindent \textbf{Multitask learning} \cite{zhang2021survey}
emerges as a solution
to learn a single model
which is shared across several tasks,
so as to achieve greater efficiency
than training dedicated models individually for each task.
Recent work~\cite{yu2020gradient} pointed out that
conflicting gradients during multitask learning
impede advancement.
To break this condition
and achieve positive interactions between tasks,
they proposed to de-conflict such gradients
by altering their directions towards a common orientation.
Our model is also constructed in a multitask learning manner,
in which the main and the auxiliary tasks are jointly optimized
during training.
However,
the auxiliary task is designed to facilitate the main task
therefore it is unsuitable to consider them in the same hierarchy.
Instead,
we propose referenced gradient calibration
by setting the main task as the reference,
and calibrating the auxiliary gradient towards it,
so as to ensure the auxiliary task
can be harmoniously trained alongside the main task,
so that it may provide supervision
for the primary model objective.

\vspace{1em}
\noindent \textbf{Test-Time model optimization}
is an emerging paradigm to tackle distribution shifts
between training and testing environments.
The key idea is to perform post-training model optimization
given the test samples during deployment.
Several recent works
\cite{wang2022continual, wang2020fully, iwasawa2021test, eastwood2021source}
proposed to optimize the model parameters
by providing proper supervision,
such as batch-norm statistics, entropy minimization, and pseudo-labeling.
Another line of work~\cite{sun2020test, liu2021tttp}
jointly trains additional self-supervised auxiliary tasks,
which are subsequently used to guide the model optimization
during testing.
This does not involve any assumptions about the output
and is therefore more generic.
It has also been applied to ReID \cite{han2022generalizable}
by considering self-supervised learning tasks
for updating BN statistics.
In this work,
we formulate \method{abbr}+ by incorporating the proposed \method{abbr} regularization
into the deployment-time optimization framework
to seek further improvement.
With the tailored auxiliary objective as the optimization supervision,
\method{abbr}+ effectively exploits the underlying target domain characteristic
and exhibits boosted performance on all the benchmarks.
\section{Methodology}
\vspace{-1em}
\noindent \paragraph{Problem Definition}
Given a labeled source domain
$\mathcal{D}_S = \left\{ (x_i, y_i) \right\}_{i \in \{1, \cdots, N\} }$
for training,
where $N$ is the number of samples,
the aim of ReID
is to learn a mapping function
parameterized by $\theta$
that projects a person image $x$
to a high-dimensional feature representation
$f_\theta$,
with the constraint that
features of the same identity have a smaller distance relative to one another.
DG ReID is more practical by assuming the non-availability
of the target domain during training,
and expects the model to be able to extract discriminative feature representations
from any target domain.
Current models are designed solely with an instance classification objective,
that can be confused by negative domain-specific information
and fall into a local optimum of the source domain.

\begin{figure*}
      \centering
      \includegraphics[width=\textwidth]{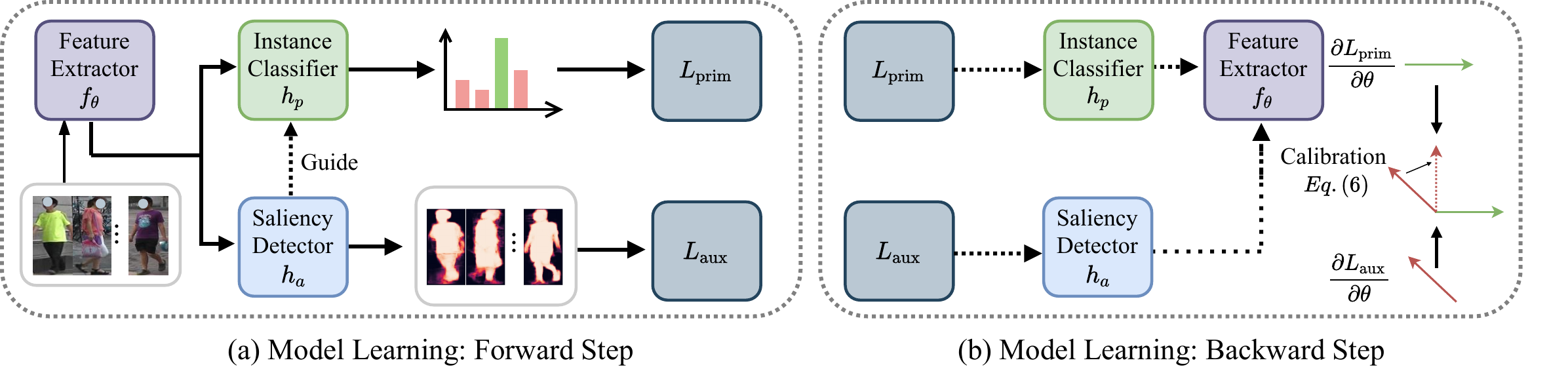}
      \caption{Overview of the proposed
            \textit{\protect\method{full}} (\protect\method{abbr}) model.
            The purpose is to derive generic feature representations
            by guiding the network to attentively focus on
            pedestrian information
            and mitigate the interference of domain-specific knowledge,
            which is achieved by the \protect \method{abbr} regularization
            of a primary classification objective
            and an auxiliary pedestrian saliency detection objective:
            (a) The auxiliary task is jointly trained
            to provide hard-coded spatial attention to the pedestrian region.
            (b) The primary task is used as a reference
            to calibrate the gradients of the auxiliary objective
            when they are conflicting.
      }
      \label{fig:framework}
      \vspace{-1em}
\end{figure*}

\subsection{Overview}
In this work,
we consider the problem of generalizing a ReID model
to any new deployment target environment
subject to unknown domain bias
between the training and the test domains, where there is no
labeled training data from the test domain.
To that end,
we propose a \textit{\method{full}} (\method{abbr}) regularization
method to enable the model to be more attentive to learning universal
identity generative information that is applicable in any domain whilst
concurrently maximizing ReID discriminative information from the domain
labeled data.
Figure~\ref{fig:framework} shows an overview of \method{abbr} in model
training with two associative steps:
(1) Guiding the ReID model to focus on discriminative pedestrian information
with an additional auxiliary task dedicated to visual saliency detection.
(2) Calibrate the gradients of the auxiliary task
when it conflicts with the primary instance classification objective.
To boost the performance,
we build \method{abbr}+
to utilize the available samples in deployment time
by minimizing the proposed auxiliary objective,
and demonstrate the plug-and-play merit of our design.

\subsection{Joint Primary-Auxiliary Objectives Learning}
The primary and auxiliary objectives are jointly trained
in a multitask learning architecture,
which is composed of a shared feature extractor $f_\theta$,
and two dedicated heads $h_p$ and $h_a$
respectively for the primary and auxiliary tasks.

\noindent \paragraph{Primary Objective: Person ReID}
Learning a strong instance classification network
is fundamentally important for training a discriminative ReID model.
Given a labeled training set $ \mathcal{D}$ = $\{ (x_i, y^{(p)}_i)\}_{ i \in \{ 1, \cdots, N \}}$,
where $x_i$ is a person image
and $y^{(p)}_i$ is the corresponding instance category label,
the primary instance classification task is trained with a softmax cross-entropy (CE) loss $\mathcal{L}_{\text{id}}$
and a triplet loss $\mathcal{L}_{\text{tri}}$:
\begin{equation}
      \mathcal{L}_{\text{id}} = - \sum_{i=1}^{N} \sum_{j=1}^{C} p_i^j \text{log} \hat{p}_i^j,
\end{equation}
where $p_i$ is one-hot vector activated at $y_i^{(p)}$,
and $\hat{p}_i^j$ is the probability for categorized into the $j$th class
that calculated from the classifier.
The additional triplet loss constrains the distance
between positive (same identity) and negative (different identities) sample pairs,
which is formulated as
\begin{equation}
      \mathcal{L}_{\text{tri}} = \sum_{i=1}^{N} [ d_p - d_n + \alpha ]_+,
\end{equation}
where $d_p$ and $d_n$ respectively denote the Euclidean distances
for the positive and negative pairs in feature space.
$\alpha$ is the margin that controls the sensitivity
and $[s]_+$ is $\text{max}(s,0)$.
The overall loss function for the primary task is as
follows:
\begin{equation}
      \mathcal{L}_\text{prim} = \mathcal{L}_{\text{id}} + \mathcal{L}_{\text{tri}}.
\end{equation}

\noindent \paragraph{Auxiliary Objective: Pedestrian Saliency Detection}
As illustrated in \cite{sun2020test}, an auxiliary task closely aligned with the primary task can substantially prompt the learning of the primary objective.  
Inspired by this, 
we formulated the auxiliary task
as pedestrian saliency detection
to perform pixel-level pedestrian localization
within the cropped pedestrian bounding boxes.
Such an auxiliary task is complementary to the primary task
by providing pixel-level hard-coded spatial attention
to guide the ReID model to focus on the pedestrian region.
Instead of exhaustively manually annotating the pedestrian region,
we benefit from the large-scale trained model \cite{zhao2019pyramid}
and perform feed-forward inference to get the weakly labelled samples.
Specifically,
given a trained saliency model $\mathcal{G}$,
we feed the sample to obtain the weak label as
$y_i^{(a)} = \mathcal{G}(x_i)$,
which is a 2D map to indicate the saliency area.
The auxiliary task
and it's essentially a regression task
in the pixel level.
To that end,
the auxiliary head $h_a$ is designed as a lightweight module
composed of cascaded 2D CNN layers
to predict the saliency map.
It is optimized by minimizing
a conventional $L1$ loss
on the predicted salient label $\hat{y}^{(a)}_k$:
\begin{equation}
      \mathcal{L}_\text{aux} = \sum_{k=1}^{N_k} | y^{(a)}_k - \hat{y}^{(a)}_k  \vert.
\end{equation}
\noindent \paragraph{Joint Multi-task Learning}
To build a joint multitask learning pipeline,
we formulate the overall objective function
by combining both $L_\text{prim}$ and $L_\text{aux}$ as
\begin{small}
      \begin{equation}
            \mathcal{L}_\text{train} = \frac{1}{N} \sum_{1}^{N} \mathcal{L}_\text{prim}(x_i,y^{(p)}_i;f_\theta,h_p) + \lambda \mathcal{L}_\text{aux}(x_i,y^{(a)}_i;f_\theta,h_a),
            \label{eq:loss_train}
      \end{equation}
\end{small}
where $\lambda$ is the balancing hyperparameter.

\begin{figure}[htbp]
      \centering
      \includegraphics[width=0.95\linewidth]{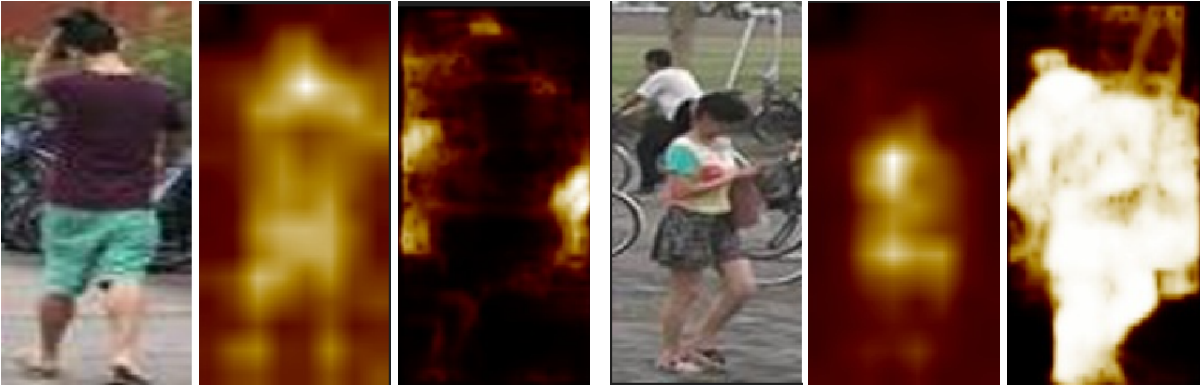}
      \caption{
            Illustration
            of the interference to the ReID objective
            when the weak saliency label is inaccurate.
            Each sample is presented with three columns:
            the input pedestrian image on the left,
            the activation from the primary ReID head in the middle,
            and the weak label for the auxiliary saliency detection head on the right.
            The gradient descent directions for the two objectives are contradictory.
      }
      \label{fig:conflict}
\end{figure}

\noindent \paragraph{Limitation:}
Despite the auxiliary objective essentially providing hard-coded spatial attention
to guide the network being focused on the salient pedestrian object,
this pipeline is intrinsically limited.
This is due to the inherent noise in the weak label of the auxiliary task that brings a detrimental impact on the primary task and distracts the shared feature extractor from focusing on the pedestrian region. This has further resulted in a divergent gradient descent direction, reflected by the conflicting gradients.
We intuitively visualize the cause of interference in Figure~\ref{fig:conflict}.
Hence, it becomes necessary 
to perform a post-operation 
that resolves the conflicts between the learning objectives.

\subsection{Association: Referenced Gradient Calibration}
During the model training,
the learnable parameter $\theta$ of the shared feature extractor $f_\theta$
is updated based on two loss gradients:
${\boldsymbol{g_p}=\frac{\partial {L_\text{prim}}}{\partial \theta}}$ from the primary objective
and $\boldsymbol{g_a}=\frac{\partial {L_\text{aux}}}{\partial \theta}$ from the auxiliary objective.
However,
when $\boldsymbol{g_p}$ and $\boldsymbol{g_a}$ are in conflict
as reflected by a negative inner product,
\ie $(\boldsymbol{g_a} \cdot \boldsymbol{g_p}) < 0$,
their joint effort cannot provide the network with
an informative direction on which to perform the gradient descent
to optimize the parameters.
Therefore,
collectively they bring significant difficulty in model convergence
and can even lead to destructive interference \cite{yu2020gradient}.

To address this fundamental limitation,
we propose to break through the dilemma
by calibrating the conflicting gradient
yield by the auxiliary objective
with that from the primary objective as a reference.
Specifically,
When $\boldsymbol{g_a}$ is conflicting with $\boldsymbol{g_p}$,
we consider $\boldsymbol{g_p}$ as a reference
and manually alter the direction of $\boldsymbol{g_a}$
by mapping it to the normal plane of $\boldsymbol{g_p}$
to get the calibrated gradient $\boldsymbol{g_a^c}$ as
\begin{equation}
      \boldsymbol{g_a^c} = \boldsymbol{g_a} - \frac{\boldsymbol{g_a} \cdot \boldsymbol{g_p}}{\| \boldsymbol{g_p} \Vert ^ 2} \boldsymbol{g_a}, \qquad \text{subject to } (\boldsymbol{g_a} \cdot \boldsymbol{g_p}) < 0,
      \label{eq:grad_calib}
\end{equation}

\vspace{0.5em}
\noindent \textbf{Remark:}
This procedure changes the direction of the conflicting gradient
to ensure it does not conflict with the primary task.
With the calibrated gradient, the model can consider the partial guidance
of the auxiliary objective, ensuring the joint effort
is non-conflicting with the primary objective.
It is effective in minimizing the side effects
caused by the inaccurate labeling of the auxiliary task
while still performing standard first-order gradient descent
to optimize the model.

\subsection{Deployment-Time Optimization}
We further formulate the \method{abbr}+
to exploit the data characteristic
of the target domain
and perform deployment time optimization
with the available samples during testing.
Considering that the proposed \method{abbr}
is composed of a shared feature encoder $f_\theta$
and two separate task heads $h_p$ and $h_a$
that are optimized jointly during model training.
When the trained model is deployed in a new environment,
given a batch of identity-unknown samples $\{ x_i^{\prime}\}_{i \in \{ 1, \cdots, B^{\prime} \}} $,
with the corresponding weakly labels $\{y_i^{\prime(a)}\}$
generated by the pre-trained saliency detection model,
the shared feature extract $f_\theta$ can be further optimized
on the auxiliary task by minimizing the following loss
\begin{equation}
      \mathcal{L}_\text{test} = \frac{1}{B} \sum_{1}^{B} \mathcal{L}_\text{aux}(x_i^{\prime}, {y_i^{\prime(a)}};f_\theta).
      \label{eq:loss_test}
\end{equation}

So that $f_\theta$ can be swiftly adapted by considering the data distribution of the new environment,
further to yield improved performance on the main task.
Note the difference from domain adaptation based methods
which assume the test sample is available during the training phase
for explicit distribution alignment,
\method{abbr}+ only requires a batch of samples with arbitrary numbers
for on-the-fly updates,
allowing it to seamlessly adapt to new data distributions.

\subsection{Model Training and Deployment}
\vspace{-1em}
\noindent \paragraph{Training stage:}
Given the formulation of the primary and auxiliary tasks,
the \method{abbr} model is designed in multitask learning architecture
and can benefit from the conventional learning supervision
by jointly minimizing the primary and auxiliary losses.
The  parameters are iteratively optimized
with the training loss (Eq.~\eqref{eq:loss_train}).
As the feature extractor parameterized by $\theta$
is shared by both the primary and auxiliary tasks,
it will be jointly updated with two gradients:
$\boldsymbol{g_p}$ for the primary task
and $\boldsymbol{g_a}$ for the auxiliary task.
To seek positive interactions between tasks,
the direction of $\boldsymbol{g_a}$ will be calibrated
only if it conflicts with $\boldsymbol{g_p}$
by Eq.~\eqref{eq:grad_calib}.
Note that the cross-entropy loss provides stronger supervision
for person classification,
therefore we use its gradients as the reference
to calibrate that of the auxiliary task.
This calibrated gradient ensures
the auxiliary task is harmoniously trained
with the primary task
by back-propagation
and thereby brings benefits
to facilitate the deployment-time optimization.
The overall training procedure is depicted in Algorithm~\ref{alg:training}.
\begin{algorithm}
      \caption{Model Training with \protect \method{abbr} regularization}
      \label{alg:training}
      \textbf{Input:}
      Labeled dataset $\mathcal{D} = \{(x_i, y^{(p)}_i)\}$ for primary task,
      weak label generator $\mathcal{G}$ for auxiliary task,
      shared feature extractor $f_\theta$,
      head modules $h_p$/$h_a$ for primary/auxiliary tasks.
      \\
      \textbf{Output:}
      Trained $f_\theta$, $h_p$ and $h_a$.
      \\
      \textbf{for} $i=1$ \textbf{to} $max\_iter$ \textbf{do}
      \\\hphantom{~~}
      Randomly sample a mini-batch $\{(x_i, y^{(p)}_i)\}_{i \in \{ 1, \cdots, N_\text{B} \} }$ from source dataset $\mathcal{D}$.
      \\\hphantom{~~}
      Generate the weak label for the auxiliary task by $\{y_i^{(a)} = \mathcal{G}(x_i) \}_{i \in \{ 1, \cdots, N_\text{B} \} }$.
      \\\hphantom{~~}
      Compute the training loss (Eq.~\eqref{eq:loss_train}) and calculate the gradients.
      \\\hphantom{~~}
      Calibrate the conflicting gradients (Eq.~\eqref{eq:grad_calib}).
      \\\hphantom{~~}
      Update the network by gradient descent.
      \\\textbf{end for}
\end{algorithm}

\vspace{-1em}
\noindent \paragraph{Deployment stage:}
To make a consistent comparison with DG ReID methods,
we can directly apply the trained \method{abbr} model for identity representation extraction.
Additionally,
the improved \method{abbr}+ model further performs deployment time optimization
during the testing stage
to mitigate the domain shift
between the training and testing domains.
Given the identity representations,
subsequent identity retrieval
is performed by a general distance metric.

\begin{figure*}[h]
      \centering
      \includegraphics[width=0.95\textwidth]{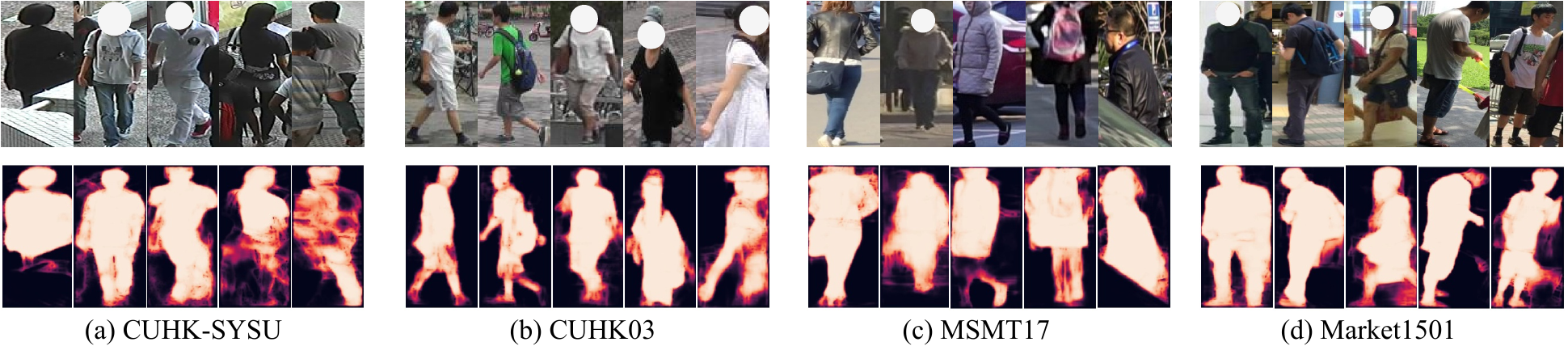}
      \caption{Example identity samples
            from different domains
            and its corresponding weak labels for the auxiliary task.
            Significant domain gaps are caused by the variation
            on nationality, illumination, viewpoints, resolution, scenario, etc.
            As complementary,
            the pedestrian saliency label can provide a guide on the most discriminative person area.}
      \label{fig:dataset_example}
      \vspace{-1em}
\end{figure*}
\section{Experiment}
\subsection{Comparison with SOTA methods}

\subsection{Experimental Settings}
\vspace{-1em}
\noindent \paragraph{Implementation Details}
We used PFAN \cite{zhao2019pyramid} as the wake label generator
for the auxiliary task.
The shared feature extractor is a ResNet50~\cite{he2016deep} pre-trained on ImageNet~\cite{deng2009imagenet}
to bootstrap the feature discrimination.
The balancing hyper-parameter in Eq.~\eqref{eq:loss_train} was set to 0.1.
The batch size was set to 64,
including 4 images for 16 randomly sampled identities.
All images were resized to $128 \times 256$.
The model was trained for 200 epochs
with the Adam optimizer \cite{kingma2014adam}.
The learning rate was set to $3.5e-4$.
The dimension of the extracted identity representation
was set to 2048.
The dimension of the saliency map is $64 \times 32$.
The learning rate for \method{abbr}+ was set to $1e-6$
and the test batch size was 200.
The post-optimization step is set to 1
for balancing performance and efficiency.
All the experiments were implemented on PyTorch~\cite{paszke2017automatic}
on a single A100 GPU.

\vspace{-1em}
\noindent \paragraph{Datasets and Evaluation Protocol}
We conducted multi-source domain generalized ReID
on a wide range of benchmarks.
including
Market1501 (M)~\cite{zheng2015scalable},
MSMT17 (MS)~\cite{wei2018person},
CUHK03 (C3)~\cite{li2014deepreid},
CUHK-SYSU (CS)~\cite{xiao2016end},
CUHK02 (C2)~\cite{li2013locally},
VIPeR~\cite{gray2008viewpoint},
PRID~\cite{hirzer2011person},
GRID~\cite{loy2010time},
and iLIDs~\cite{zheng2009associating}.
We evaluated the performance of \method{abbr}
on the four small-scale datasets
following the traditional setting~\cite{song2019generalizable,jin2020feature,bai2021person30k,zhang2022adaptive}.
We also performed leave-one-out evaluations
by using three datasets for training and the remaining for the test
\cite{zhao2021learning,choi2021meta,liao2020interpretable}.
Note that the CUHK-SYSU is only for training
given all the images are captured by the same camera.
To learn a discriminative model benefits from diverse identities,
all the identities
regardless of the original train/test splits,
were used for training.
We
adopted Mean average precision (mAP)
and Rank-1 of CMC as the evaluation metrics.

\begin{table*}[h]
      \centering
      \caption{Comparison with the SOTA methods on traditional evaluation protocol.
            The best results are shown in \best{red}
            and the second-best results are shown in \scnd{blue}.
      }
            \begin{tabular}{l|l|cc|cc|cc|cc|cc}
\hline
 &  & \multicolumn{2}{c|}{PRID} & \multicolumn{2}{c|}{GRID} & \multicolumn{2}{c|}{VIPeR} & \multicolumn{2}{c|}{iLIDs} & \multicolumn{2}{c}{Average}\\ 
\cline{3-12}
\multirow{-2}{*}{Source} & \multirow{-2}{*}{Method} & mAP & R1 & mAP & R1 & mAP & R1 & mAP & R1 & mAP & R1\\ 
\hline
 & DIMN~\cite{song2019generalizable} & 52.0 & 39.2 & 41.1 & 29.3 & 60.1 & 51.2 & 78.4 & 70.2 & 57.9 & 47.5\\ 
 & SNR~\cite{jin2020style} & 66.5 & 52.1 & 47.7 & 40.2 & 61.3 & 52.9 & 89.9 & 84.1 & 66.3 & 57.3\\ 
\multirow{-3}{*}{\makecell[l]{M+D+C2\\ +C3+CS}} & DMG-Net~\cite{bai2021person30k} & 68.4 & 60.6 & 56.6 & 51.0 & 60.4 & 53.9 & 83.9 & 79.3 & 67.3 & 61.2\\ 
\hline
 & M3L~\cite{zhao2021learning} & 64.3 & 53.1 & 55.0 & 44.4 & 66.2 & 57.5 & 81.5 & 74.0 & 66.8 & 57.2\\ 
 & MetaBIN~\cite{choi2021meta} & 70.8 & 61.2 & 57.9 & 50.2 & 64.3 & 55.9 & 82.7 & 74.7 & 68.9 & 60.5\\ 
 & ACL~\cite{zhang2022adaptive} & 73.5 & 63.0 & 65.7 & 55.2 & 75.1 & 66.4 & 86.5 & 81.8 & 75.2 & 66.6\\ 
 & META~\cite{xu2022mimic} & 71.7 & 61.9 & 60.1 & 52.4 & 68.4 & 61.5 & 83.5 & 79.2 & 70.9 & 63.8\\ 
\cline{2-12}
 & \method{abbr} (Ours) & \scnd{{74.0}} & \scnd{{65.6}} & \scnd{{67.2}} & \scnd{{56.3}} & \scnd{{76.6}} & \scnd{{66.7}} & \scnd{{87.1}} & \scnd{{83.1}} & \scnd{{76.2}} & \scnd{{67.9}}\\ 
\multirow{-6}{*}{\makecell[l]{M+C2+\\ C3+CS}} & \method{abbr}+ (Ours) & \best{{75.1}} & \best{{66.5}} & \best{{67.8}} & \best{{56.9}} & \best{{77.2}} & \best{{67.7}} & \best{{88.0}} & \best{{83.9}} & \best{{77.0}} & \best{{68.8}}\\ 
\hline
\end{tabular}
      \label{tab:small}
\end{table*}

\begin{table*}
      \centering
      \vspace{-1em}
      \caption{Comparison with the SOTA methods on large-scale evaluation protocol.
            The best results are shown in \best{red}
            and the second-best results are shown in \scnd{blue}.
      }
            \begin{tabular}{l|l|cc|cc|cc|cc}
\hline
 & \multicolumn{1}{c|}{} & \multicolumn{2}{c|}{M+MS+CS→C3} & \multicolumn{2}{c|}{M+CS+C3→MS} & \multicolumn{2}{c|}{MS+CS+C3→M} & \multicolumn{2}{c}{Average}\\ 
\cline{3-10}
\multirow{-2}{*}{Method} & \multicolumn{1}{l|}{\multirow{-2}{*}{Reference}} & mAP & R1 & mAP & R1 & mAP & R1 & mAP & R1\\ 
\hline
SNR~\cite{jin2020style} & CVPR2020 & 17.5 & 17.1 & 7.7 & 22.0 & 52.4 & 77.8 & 25.9 & 39.0\\ 
QAConv$_{50}$~\cite{liao2020interpretable} & ECCV2020 & 32.9 & 33.3 & 17.6 & 46.6 & 66.5 & 85.0 & 39.0 & 55.0\\ 
M$^3$L~\cite{zhao2021learning} & CVPR2021 & 35.7 & 36.5 & 17.4 & 38.6 & 62.4 & 82.7 & 38.5 & 52.6\\ 
MetaBIN~\cite{choi2021meta} & CVPR2021 & 43.0 & 43.1 & 18.8 & 41.2 & 67.2 & 84.5 & 43.0 & 56.3\\ 
ACL~\cite{zhang2022adaptive} & ECCV2022 & 49.4 & 50.1 & 21.7 & 47.3 & 76.8 & 90.6 & 49.3 & 62.7\\ 
META~\cite{xu2022mimic} & ECCV2022 & 47.1 & 46.2 & 24.4 & \scnd{{52.1}} & 76.5 & 90.5 & 49.3 & 62.9\\ 
\hline
\method{abbr} & Ours & \scnd{{49.8}} & \scnd{{50.5}} & \scnd{{25.1}} & 51.5 & \scnd{{77.1}} & \scnd{{90.8}} & \scnd{{50.7}} & \scnd{{64.3}}\\ 
\method{abbr}+ & Ours & \best{{50.3}} & \best{{50.9}} & \best{{26.0}} & \best{{52.8}} & \best{{77.9}} & \best{{91.4}} & \best{{51.4}} & \best{{65.0}}\\ 
\hline
\end{tabular}
      \label{tab:sota}
\end{table*}

We compared the proposed \method{abbr}
against several recent SOTA methods,
and the comparison results are shown in
Table~\ref{tab:small} and
Table~\ref{tab:sota}.
Under a fair comparison with existing DG ReID methods, 
the \method{abbr} model
outperforms all the competing methods by a significant margin
on both the traditional setting and the large-scale settings
across all the evaluation metrics.
It shows a clear advantage over the recent SOTA methods.
Notably, even trained with fewer datasets compared with~\cite{song2019generalizable,jin2020style,bai2021person30k},
the proposed method is still able to extract discriminative features
for identity matching.
Besides, 
we extended our analysis to include the results from the
test-time optimization variant, \method{abbr}+,
which notably improves \method{abbr} consistently 
across all benchmarks.
These results provide additional evidence
on the effectiveness of the associative learning strategy,
where the auxiliary task can promote the primary ReID objective
during test time given the absence of identity labels.

\subsection{Ablation Studies}
\label{sec:abl}

\begin{table}[h]
  \centering
  \caption{Effects on mAP (\%) value of the proposed modules. 
  Aux: auxiliary objective.
  GC: gradient calibration.
  DTO: deployment-time optimization.
  }
  \resizebox{0.9\linewidth}{!}{
    \begin{tabular}{ccccccc}
    \hline
    Aux   & GC    & DTO &   C3 & MS & \multicolumn{1}{c}{M} & Average \\
    \hline
    \xmark & \xmark & \xmark & 42.8  & 20.5  & 73.1  & 45.5 \\
    \cmark & \xmark & \xmark & 44.8  & 20.9  & 73.5  & 46.4 \\
    \cmark & \xmark & \cmark & 47.0  & 23.1  & 75.2  & 48.4 \\
    \cmark & \cmark & \xmark & 49.8  & 25.1  & 77.1  & 50.7 \\
\hline    \cmark & \cmark & \cmark & \best{{50.3}}  & \best{{26.0}}  & \best{{77.9}}  & \best{{51.4}} \\
    \hline
    \end{tabular}%
    }
    \vspace{-1em}
  \label{tab:comp_abl}%
\end{table}%

\noindent \paragraph{Component Analysis}
We investigated the effects of different components
in \method{abbr} model design
to study their individual contributions.
The baseline model is a ResNet50 pre-trained on ImageNet.
The comparison results are shown in Table~\ref{tab:comp_abl},
from which we can observe that
the auxiliary objective
and the gradient calibration strategies
can consistently improve performance.
With further deployment-time optimization,
our model can be advanced by benefiting from mining the data characteristics
in the target domain.
It is notable that the variant without gradient calibration can always benefit more from that post-optimization
compared with the \method{abbr}+ model,
This further illustrates that the referenced calibration mechanism
has already enabled the \method{abbr} model to be more attentive 
to the domain-invariant pedestrian region, and therefore it relies less on on-the-fly optimization.

\begin{table}[h]
  \centering
  \caption{Effects on mAP (\%s) of update iterations during deployment optimization.}
      \resizebox{0.89\linewidth}{!}{
    \begin{tabularx}{0.45\textwidth}{l@{\extracolsep{\fill}}cccccc}
    \hline
    Dataset & 0     & 1     & 2     & 3     & 4     \\
    \hline
    C3 & 49.8  & 50.3  & 50.5  & 50.6  & 50.3  \\
    MS & 25.1  & 26.0  & 26.5  & 26.0  & 25.0  \\
    M & 77.1  &  77.9 & 77.5  & 77.0  & 76.2  \\
    \hline
    Avg. & 50.7  & 51.4  & 51.5  & 51.2  & 50.5  \\
    \hline
    \end{tabularx}%
    }
  \label{tab:update_iter}%
 \vspace{-1em} 
\end{table}%

\begin{figure}[h]
      \centering
      \resizebox{\linewidth}{!}{
            \includegraphics{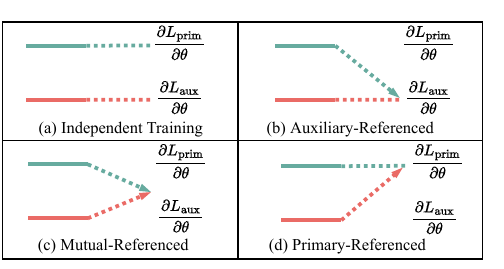}
      }
      \vspace{-1em}
      \caption{Illustration of different gradient calibration designs.
            (a) No gradient calibration as~\cite{song2018mask}.
            (b) Gradients of the primary objective are calibrated with the auxiliary objective as a reference.
            (c) Gradients are calibrated in relation to each other as a reference, 
            as designed in~\cite{yu2020gradient}.
            (d) Gradients of the auxiliary objective are calibrated with the primary objective as a reference.}
      \label{fig:abl_calib}
      \vspace{-1em}
\end{figure}

\vspace{-1em}
\noindent \paragraph{Gradient Calibration Designs}
We adopted a primary-referenced design
for the gradient calibration
between the primary and auxiliary objectives.
This was based on the fact that
the primary instance classification objective
provides stronger supervision to identify pedestrians,
while the auxiliary objective is to guide the instance classifier
to attentively focus on the pedestrian area
and ignore the domain-specific interference.
It's weakly labeled and therefore is intricately noisy which
can lead to a negative influence on the primary objective,
reflected by the conflicting gradient.
We examined the effect of the calibration design
by additionally testing three more formulations
as demonstrated in Figure~\ref{fig:abl_calib}.
Table~\ref{tab:calib} shows
the auxiliary-referenced design yielded the worst performance,
given the gradients of the auxiliary objective
is noisy and unreliable,
using it as the reference
is harmful to the learning of the primary objective.
By contrast,
the mutually referenced calibration design  includes
the primary gradients as referenced on top of the auxiliary-referenced design,
which alleviates the fallout caused by the gradient destruction,
despite it's still inferior to the baseline.
In comparison,
the primary-referenced design consistently obtained improved performance
which supports the design of the proposed
primary referenced gradient calibration.

\vspace{-1em}
\begin{table}[h]
      \centering
      \caption{Comparison of different gradient calibration designs by mAP (\%).
            Refer to Figure~\ref{fig:abl_calib} for the corresponding design.
      }
      \resizebox{0.95\linewidth}{!}{
            \begin{tabularx}{0.45\textwidth}{l@{\extracolsep{\fill}}cccc}
\hline
Design & C3 & MS & M & Avg. \\ 
\hline
a & 44.8 & 20.9 & 73.5 & 46.4\\ 
b & 44.1 & 21.7 & 74.7 & 46.8\\ 
c & 47.3 & 23.0 & 75.3 & 48.5\\ 
\hline
d & \best{{49.8}} & \best{{25.1}} & \best{{77.1}} & \best{{50.7}}\\ 
\hline
\end{tabularx}
      }
      \label{tab:calib}
      \vspace{-1em}
\end{table}

\noindent \paragraph{Update iterations for deployment-time optimization}
We analyzed the influence of update iterations for optimizing the model with all test samples in deployment time. Ablating with iterations from 0 to 4 (Table~\ref{tab:update_iter}), we noted consistent performance improvement by updating the model at the initial steps. This is attributed to the auxiliary objective guiding swift adaptation to the test domain.  This improvement is attributed to the auxiliary objective facilitating rapid adaptation to the test domain. However, excessive updates result in a model forgetting issue by overwhelming the extractor with the auxiliary. Notably, deployment-time optimization is more effective for target datasets (\ie MSMT17) with larger domain shifts, which further proves that target-aware updates that mitigate domain shifts more effectively. Balancing efficiency and effectiveness, PAOA+ adopts single-step updates across all datasets to attain the global optimal solution.

\begin{figure}[h]
      \centering
      \resizebox{\linewidth}{!}{
            \includegraphics{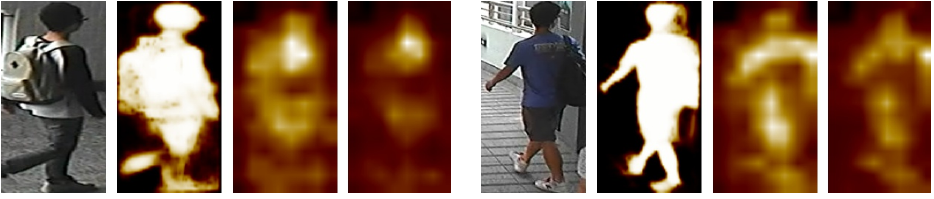}
      }
      \caption{Visualization of activation maps.
            For each pedestrian image, the four columns from left to right are:
            (1) Person image,
            (2) Weak label for auxiliary objective,
            (3) Activation map from the proposed \protect \method{abbr} model,
            (4) Activation map from the baseline.
            The proposed \protect \method{abbr} helps the model be more attentive
            on the pedestrian region to learning domain-invariant representation.
      }
      \label{fig:vis}
      \vspace{-1em}
\end{figure}

\begin{figure}[ht]
      \centering
      \resizebox{\linewidth}{!}{
            \includegraphics{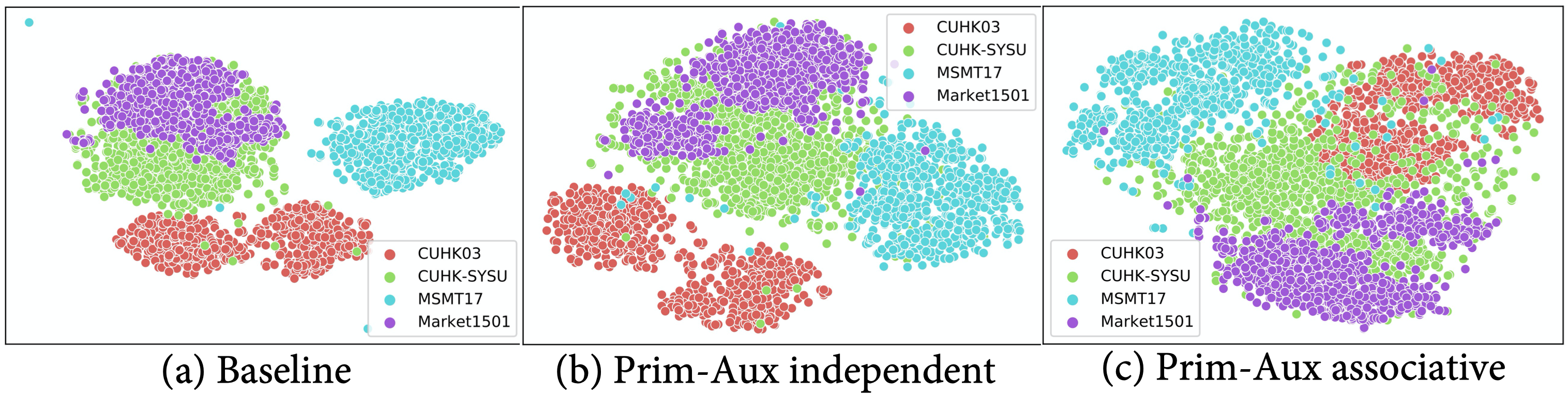}
      }
      \caption{TSNE visualization on extracted features.
            200 samples were randomly sampled from each domain.
            Learning with jointly the primary and auxiliary objectives
            can condense the feature distribution.
            The proposed model which associates the primary and auxiliary objectives can derive a more compact feature space.
      }
      \label{fig:tsne}
\end{figure}

\vspace{-1em}
\noindent \paragraph{Visualization}
We visualized the pedestrian images and the model activation maps
to intuitively illustrate the effectiveness of \method{abbr}.
We took the feature map of the final convolutional layer ($4th$ layer)
as the activation map,
and compared the baseline model
with the proposed \method{abbr}.
As can be observed in Figure~\ref{fig:vis},
the \method{abbr} model can accurately be attentive to the pedestrian area,
while the baseline model is partially focus
and some discriminative areas are missed.
This is benefited from the auxiliary objective,
as shown in the second column,
which provides assistive supervision
on instance classification learning.
Therefore,
\method{abbr} model can extract discriminative yet generic identity representation
for ReID.
To also visualized the TSNE distribution of the extracted feature representations in Figure~\ref{fig:tsne}.
The target domain is Market1501
and the model was trained with other three source domains.
Training independently with the auxiliary objective
can condense the feature space
compared with the baseline,
however it's still prone to domain shift,
especially for CUHK03.
As a comparison,
the proposed \protect \method{abbr} can significantly reduce domain shifts
with a much more compact feature space.

\section{Conclusions}
In this work,
we introduced a novel \textit{\method{full}} (\method{abbr}) regularization
to learn a generalizable ReID model
for extracting domain-unbiased representations
more generalizable to unseen novel domains for person ReID.
\method{abbr} encourages the model to get rid of the interference
of domain-specific knowledge
and to learn from discriminative pedestrian information
by the association of learning an auxiliary pedestrian detection objective
with a primary instance classification objective.
To mitigate the fallout caused by the noisy auxiliary labels,
we further derive a referenced-gradient calibration strategy
to alter the gradient of the auxiliary object
when it's conflicting with the primary object.
The \method{abbr} framework is task-agnostic,
making it readily adaptable to other tasks 
through the incorporation of a close auxiliary task 
and a shared learning module.

{\small
\bibliographystyle{ieee_fullname}
\bibliography{ref}

\begin{thebibliography}{10}\itemsep=-1pt

\bibitem{ang2021dex}
Eugene~PW Ang, Lin Shan, and Alex~C Kot.
\newblock Dex: Domain embedding expansion for generalized person re-identification.
\newblock In {\em BMVC}, 2021.

\bibitem{bai2021person30k}
Yan Bai, Jile Jiao, Wang Ce, Jun Liu, Yihang Lou, Xuetao Feng, and Ling-Yu Duan.
\newblock Person30k: A dual-meta generalization network for person re-identification.
\newblock In {\em CVPR}, 2021.

\bibitem{borji2019salient}
Ali Borji, Ming-Ming Cheng, Qibin Hou, Huaizu Jiang, and Jia Li.
\newblock Salient object detection: A survey.
\newblock {\em Computational visual media}, 5(2), 2019.

\bibitem{choi2021meta}
Seokeon Choi, Taekyung Kim, Minki Jeong, Hyoungseob Park, and Changick Kim.
\newblock Meta batch-instance normalization for generalizable person re-identification.
\newblock In {\em CVPR}, 2021.

\bibitem{dai2021generalizable}
Yongxing Dai, Xiaotong Li, Jun Liu, Zekun Tong, and Ling-Yu Duan.
\newblock Generalizable person re-identification with relevance-aware mixture of experts.
\newblock In {\em CVPR}, 2021.

\bibitem{deng2009imagenet}
Jia Deng, Wei Dong, Richard Socher, Li-Jia Li, Kai Li, and Li Fei-Fei.
\newblock Imagenet: A large-scale hierarchical image database.
\newblock In {\em CVPR}, 2009.

\bibitem{eastwood2021source}
Cian Eastwood, Ian Mason, Christopher~KI Williams, and Bernhard Sch{\"o}lkopf.
\newblock Source-free adaptation to measurement shift via bottom-up feature restoration.
\newblock In {\em ICLR}, 2022.

\bibitem{gray2008viewpoint}
Douglas Gray and Hai Tao.
\newblock Viewpoint invariant pedestrian recognition with an ensemble of localized features.
\newblock In {\em ECCV}, 2008.

\bibitem{han2022generalizable}
Ke Han, Chenyang Si, Yan Huang, Liang Wang, and Tieniu Tan.
\newblock Generalizable person re-identification via self-supervised batch norm test-time adaption.
\newblock In {\em AAAI}, 2022.

\bibitem{he2016deep}
Kaiming He, Xiangyu Zhang, Shaoqing Ren, and Jian Sun.
\newblock Deep residual learning for image recognition.
\newblock In {\em CVPR}, 2016.

\bibitem{he2020guided}
Lingxiao He and Wu Liu.
\newblock Guided saliency feature learning for person re-identification in crowded scenes.
\newblock In {\em ECCV}, 2020.

\bibitem{hirzer2011person}
Martin Hirzer, Csaba Beleznai, Peter~M Roth, and Horst Bischof.
\newblock Person re-identification by descriptive and discriminative classification.
\newblock In {\em Scandinavian conference on Image analysis}, 2011.

\bibitem{iwasawa2021test}
Yusuke Iwasawa and Yutaka Matsuo.
\newblock Test-time classifier adjustment module for model-agnostic domain generalization.
\newblock {\em NeurIps}, 2021.

\bibitem{jia2019frustratingly}
Jieru Jia, Qiuqi Ruan, and Timothy~M Hospedales.
\newblock Frustratingly easy person re-identification: Generalizing person re-id in practice.
\newblock In {\em BMVC}, 2019.

\bibitem{jin2020feature}
Xin Jin, Cuiling Lan, Wenjun Zeng, and Zhibo Chen.
\newblock Feature alignment and restoration for domain generalization and adaptation.
\newblock {\em arXiv}, 2020.

\bibitem{jin2020style}
Xin Jin, Cuiling Lan, Wenjun Zeng, Zhibo Chen, and Li Zhang.
\newblock Style normalization and restitution for generalizable person re-identification.
\newblock In {\em CVPR}, 2020.

\bibitem{kingma2014adam}
Diederik~P Kingma and Jimmy Ba.
\newblock Adam: A method for stochastic optimization.
\newblock In {\em ICLR}, 2015.

\bibitem{li2021local}
Qilei Li, Jiabo Huang, and Shaogang Gong.
\newblock Local-global associative frame assemble in video re-id.
\newblock {\em BMVC}, 2021.

\bibitem{li2013locally}
Wei Li and Xiaogang Wang.
\newblock Locally aligned feature transforms across views.
\newblock In {\em CVPR}, 2013.

\bibitem{li2014deepreid}
Wei Li, Rui Zhao, Tong Xiao, and Xiaogang Wang.
\newblock Deepreid: Deep filter pairing neural network for person re-identification.
\newblock In {\em CVPR}, 2014.

\bibitem{li2018harmonious}
Wei Li, Xiatian Zhu, and Shaogang Gong.
\newblock Harmonious attention network for person re-identification.
\newblock In {\em CVPR}, 2018.

\bibitem{liao2020interpretable}
Shengcai Liao and Ling Shao.
\newblock Interpretable and generalizable person re-identification with query-adaptive convolution and temporal lifting.
\newblock In {\em ECCV}, 2020.

\bibitem{liu2021tttp}
Yuejiang Liu, Parth Kothari, Bastien van Delft, Baptiste Bellot-Gurlet, Taylor Mordan, and Alexandre Alahi.
\newblock Ttt++: When does self-supervised test-time training fail or thrive?
\newblock In {\em NeuIPS}, 2021.

\bibitem{loy2010time}
Chen~Change Loy, Tao Xiang, and Shaogang Gong.
\newblock Time-delayed correlation analysis for multi-camera activity understanding.
\newblock {\em IJCV}, 2010.

\bibitem{luo2020generalizing}
Chuanchen Luo, Chunfeng Song, and Zhaoxiang Zhang.
\newblock Generalizing person re-identification by camera-aware invariance learning and cross-domain mixup.
\newblock In {\em ECCV}, 2020.

\bibitem{mahajan2021domain}
Divyat Mahajan, Shruti Tople, and Amit Sharma.
\newblock Domain generalization using causal matching.
\newblock In {\em ICML}, 2021.

\bibitem{paszke2017automatic}
Adam Paszke, Sam Gross, Soumith Chintala, Gregory Chanan, Edward Yang, Zachary DeVito, Zeming Lin, Alban Desmaison, Luca Antiga, and Adam Lerer.
\newblock Automatic differentiation in pytorch.
\newblock In {\em NIPS-W}, 2017.

\bibitem{sener2018multi}
Ozan Sener and Vladlen Koltun.
\newblock Multi-task learning as multi-objective optimization.
\newblock {\em Advances in neural information processing systems}, 31, 2018.

\bibitem{song2018mask}
Chunfeng Song, Yan Huang, Wanli Ouyang, and Liang Wang.
\newblock Mask-guided contrastive attention model for person re-identification.
\newblock In {\em CVPR}, 2018.

\bibitem{song2019generalizable}
Jifei Song, Yongxin Yang, Yi-Zhe Song, Tao Xiang, and Timothy~M Hospedales.
\newblock Generalizable person re-identification by domain-invariant mapping network.
\newblock In {\em CVPR}, 2019.

\bibitem{sun2020test}
Yu Sun, Xiaolong Wang, Zhuang Liu, John Miller, Alexei Efros, and Moritz Hardt.
\newblock Test-time training with self-supervision for generalization under distribution shifts.
\newblock In {\em ICML}, 2020.

\bibitem{wang2020fully}
Dequan Wang, Evan Shelhamer, Shaoteng Liu, Bruno~A Olshausen, and Trevor Darrell.
\newblock Fully test-time adaptation by entropy minimization.
\newblock In {\em ICLR}, 2021.

\bibitem{wang2022continual}
Qin Wang, Olga Fink, Luc Van~Gool, and Dengxin Dai.
\newblock Continual test-time domain adaptation.
\newblock In {\em CVPR}, 2022.

\bibitem{wei2018person}
Longhui Wei, Shiliang Zhang, Wen Gao, and Qi Tian.
\newblock Person transfer gan to bridge domain gap for person re-identification.
\newblock In {\em CVPR}, 2018.

\bibitem{xiao2016end}
Tong Xiao, Shuang Li, Bochao Wang, Liang Lin, and Xiaogang Wang.
\newblock End-to-end deep learning for person search.
\newblock {\em arXiv}, 2016.

\bibitem{xu2022mimic}
Boqiang Xu, Jian Liang, Lingxiao He, and Zhenan Sun.
\newblock Mimic embedding via adaptive aggregation: Learning generalizable person re-identification.
\newblock In {\em Computer Vision--ECCV 2022: 17th European Conference, Tel Aviv, Israel, October 23--27, 2022, Proceedings, Part XIV}, pages 372--388. Springer, 2022.

\bibitem{yu2020gradient}
Tianhe Yu, Saurabh Kumar, Abhishek Gupta, Sergey Levine, Karol Hausman, and Chelsea Finn.
\newblock Gradient surgery for multi-task learning.
\newblock {\em NeurIps}, 2020.

\bibitem{zhang2022adaptive}
Pengyi Zhang, Huanzhang Dou, Yunlong Yu, and Xi Li.
\newblock Adaptive cross-domain learning for generalizable person re-identification.
\newblock In {\em ECCV}, 2022.

\bibitem{zhang2021survey}
Yu Zhang and Qiang Yang.
\newblock A survey on multi-task learning.
\newblock {\em IEEE Transactions on Knowledge and Data Engineering}, 2021.

\bibitem{zhang2020relation}
Zhizheng Zhang, Cuiling Lan, Wenjun Zeng, Xin Jin, and Zhibo Chen.
\newblock Relation-aware global attention for person re-identification.
\newblock In {\em CVPR}, 2020.

\bibitem{zhao2019pyramid}
Ting Zhao and Xiangqian Wu.
\newblock Pyramid feature attention network for saliency detection.
\newblock In {\em CVPR}, 2019.

\bibitem{zhao2021learning}
Yuyang Zhao, Zhun Zhong, Fengxiang Yang, Zhiming Luo, Yaojin Lin, Shaozi Li, and Nicu Sebe.
\newblock Learning to generalize unseen domains via memory-based multi-source meta-learning for person re-identification.
\newblock In {\em CVPR}, 2021.

\bibitem{zheng2015scalable}
Liang Zheng, Liyue Shen, Lu Tian, Shengjin Wang, Jingdong Wang, and Qi Tian.
\newblock Scalable person re-identification: A benchmark.
\newblock In {\em ICCV}, 2015.

\bibitem{zheng2009associating}
Wei-Shi Zheng, Shaogang Gong, and Tao Xiang.
\newblock Associating groups of people.
\newblock In {\em BMVC}, 2009.

\bibitem{zheng2019joint}
Zhedong Zheng, Xiaodong Yang, Zhiding Yu, Liang Zheng, Yi Yang, and Jan Kautz.
\newblock Joint discriminative and generative learning for person re-identification.
\newblock In {\em CVPR}, 2019.

\bibitem{zhou2021domain}
Kaiyang Zhou, Ziwei Liu, Yu Qiao, Tao Xiang, and Chen~Change Loy.
\newblock Domain generalization in vision: A survey.
\newblock {\em arXiv}, 2021.

\bibitem{zhou2020learning}
Kaiyang Zhou, Yongxin Yang, Timothy Hospedales, and Tao Xiang.
\newblock Learning to generate novel domains for domain generalization.
\newblock In {\em ECCV}, 2020.

\bibitem{zhu2020identity}
Kuan Zhu, Haiyun Guo, Zhiwei Liu, Ming Tang, and Jinqiao Wang.
\newblock Identity-guided human semantic parsing for person re-identification.
\newblock In {\em ECCV}, 2020.

\bibitem{zhuang2021camera}
Zijie Zhuang, Longhui Wei, Lingxi Xie, Haizhou Ai, and Qi Tian.
\newblock Camera-based batch normalization: an effective distribution alignment method for person re-identification.
\newblock {\em IEEE Transactions on Circuits and Systems for Video Technology}, 32(1), 2021.

\bibitem{zhuang2020rethinking}
Zijie Zhuang, Longhui Wei, Lingxi Xie, Tianyu Zhang, Hengheng Zhang, Haozhe Wu, Haizhou Ai, and Qi Tian.
\newblock Rethinking the distribution gap of person re-identification with camera-based batch normalization.
\newblock In {\em ECCV}, 2020.

\end{thebibliography}
}
\end{document}